\documentclass[lettersize,journal]{IEEEtran}
\usepackage{amsmath,amssymb,amsfonts}
\usepackage[ruled,vlined]{algorithm2e}
\usepackage{array}
\usepackage{graphicx}
\usepackage{booktabs}
\usepackage{tabularx}
\usepackage{subcaption}
\usepackage{textcomp}
\usepackage{stfloats}
\usepackage{url}
\usepackage{verbatim}
\usepackage{cite}
\usepackage[colorlinks=true,allcolors=blue]{hyperref}
\begin{document}

\title{Hybrid Quantum-Classical Spatiotemporal Forecasting for 3D Cloud Fields}

\author{Fu Wang,~\IEEEmembership{Member,~IEEE,}
Qifeng Lu,
Xinyu Long,
Meng Zhang, 
Xiaofei Yang,~\IEEEmembership{Member,~IEEE,} 
Weijia Cao, ~\IEEEmembership{Member,~IEEE,}
Xiaowen Chu, ~\IEEEmembership{Fellow,~IEEE,}
\thanks{This work was supported in part by the Innovation Fund of China Meteorological Administration, the National Natural Science Foundation of China under Grant U2242212 and 42471437. (Corresponding author: Qifeng Lu, Xiaofei Yang and Xiaowen Chu). 
Fu WANG, Qifeng LU are with CMA Earth system Modeling and Prediction Center (CEMC) and State Key Laboratory of Severe Weather (LaSW), Beijing, 100081, China (e-mail: wangfu@cma.cn, luqf@cma.cn).
Xinyu Long and Meng Zhang are with the School of Computer Science and Engineering, Sichuan University of Science and Engineering, Yibin City, 643000, China (e-mail: yangchi\underline{~}001@foxmail.com,lxf1969@suse.edu.cn). 
Xiaofei Yang is with the School of Electronic and Communication Engineering, Guangzhou University, Guangzhou, 510006, China (email:xiaofeiyang@gzhu.edu.cn ).
Weijia Cao is with Aerospace Information Research Institute, Chinese Academy of Sciences, Beijing, 100094, China (e-mail: caowj@aircas.ac.cn)
Xiaowen Chu is wih the Data Science and Analytics, Hong Kong University of Science and Technology(Guangzhou), Guangzhou, 511453, China (e-mail:xwchu@hkust-gz.edu.cn)}}%

\markboth{IEEE Journal Template}%
{Fu Wang, \MakeLowercase{\textit{et al.}}: Quantum-Enhanced Neural Operators for 3D Cloud Spatio-temporal Forecasting}

\maketitle

\begin{abstract}
Accurate forecasting of three-dimensional (3D) cloud fields is important for atmospheric analysis and short-range numerical weather prediction, yet it remains challenging because cloud evolution involves cross-layer interactions, nonlocal dependencies, and multiscale spatiotemporal dynamics. Existing spatiotemporal prediction models based on convolutions, recurrence, or attention often rely on locality-biased representations and therefore struggle to preserve fine cloud structures in volumetric forecasting tasks. To address this issue, we propose QENO, a hybrid quantum-inspired spatiotemporal forecasting framework for 3D cloud fields. The proposed architecture consists of four components: a classical spatiotemporal encoder for compact latent representation, a topology-aware quantum enhancement block for modeling nonlocal couplings in latent space, a dynamic fusion temporal unit for integrating measurement-derived quantum features with recurrent memory, and a decoder for reconstructing future cloud volumes. Experiments on CMA-MESO 3D cloud fields show that QENO consistently outperforms representative baselines, including ConvLSTM, PredRNN++, Earthformer, TAU, and SimVP variants, in terms of MSE, MAE, RMSE, SSIM, and threshold-based detection metrics. In particular, QENO achieves an MSE of 0.2038, an RMSE of 0.4514, and an SSIM of 0.6291, while also maintaining a compact parameter budget. These results indicate that topology-aware hybrid quantum-classical feature modeling is a promising direction for 3D cloud structure forecasting and atmospheric Earth observation data analysis.
\end{abstract}

\begin{IEEEkeywords}
Remote sensing, numerical weather prediction, artificial intelligence, 3D cloud forecasting, quantum-enhanced learning
\end{IEEEkeywords}
\section{Introduction}

\IEEEPARstart{T}{hree}-dimensional (3D) cloud structure plays a fundamental role in atmospheric analysis, numerical weather prediction, and downstream precipitation forecasting. Compared with scalar meteorological variables, cloud fields exhibit stronger structural complexity because their evolution is governed by multiscale transport, cross-layer coupling, and nonlinear physical processes. As a result, accurate forecasting of cloud morphology remains substantially more difficult than forecasting lower-dimensional atmospheric variables. In Earth observation applications, reliable cloud-field prediction is also closely related to the interpretation of observation-derived atmospheric information, the use of forecast data, and the integration of simulated products in weather analysis workflows.

Traditional numerical weather prediction (NWP) systems model cloud evolution through microphysical parameterizations and data assimilation schemes. Although these approaches remain essential in operational forecasting, they often introduce considerable uncertainty in the representation of vertical cloud structure, especially under complex convective conditions. Recent advances in data-driven forecasting have significantly improved the prediction of core meteorological variables, but 3D cloud-field forecasting is still relatively underexplored. Many existing studies focus on simplified cloud descriptors, such as cloud-top properties or bulk cloud content, rather than directly modeling the full spatiotemporal evolution of volumetric cloud structure.

Deep spatiotemporal forecasting models, including recurrent, convolutional, and attention-based architectures, have shown strong capability in video prediction and environmental forecasting. However, when applied to 3D cloud fields, these models face two limitations. First, most of them are dominated by local receptive fields or locality-constrained attention mechanisms, which makes it difficult to capture long-range dependencies across distant cloud regions and vertical layers. Second, multiscale cloud dynamics involve highly coupled nonlinear interactions, which are not always well preserved by purely classical latent representations. These limitations often lead to blurred boundaries, weakened structural consistency, and loss of fine-scale cloud morphology in future predictions.

Motivated by these observations, we explore a hybrid quantum-classical strategy for 3D cloud forecasting. Our objective is not to claim practical quantum advantage, but to investigate whether shallow quantum-inspired latent transformations can provide a compact mechanism for modeling nonlocal correlations and cross-layer interactions in volumetric cloud fields. Based on this idea, we develop QENO, a topology-aware hybrid forecasting framework that couples a classical spatiotemporal encoder-decoder backbone with a quantum enhancement module and a dynamic temporal fusion mechanism. The quantum branch operates on compact latent features, transforms them through parameterized circuits guided by topology-aware entanglement patterns, and feeds the resulting measurement features back into temporal state updates.

The main contributions of this work are summarized as follows.
\begin{enumerate}
\item We propose a hybrid quantum-inspired spatiotemporal forecasting framework for 3D cloud fields, in which a topology-aware quantum enhancement block is introduced to model compact nonlocal interactions in latent space.
\item We design a dynamic fusion temporal unit that integrates measurement-derived quantum features with classical recurrent memory, enabling the model to better capture cross-layer and multiscale cloud evolution.
\item We evaluate the proposed method on operational CMA-MESO cloud fields and compare it with several representative spatiotemporal baselines, demonstrating consistent improvements in both regression accuracy and structural fidelity.
\end{enumerate}
\begin{figure}
  \centering 
  \includegraphics[width=0.55\textwidth]{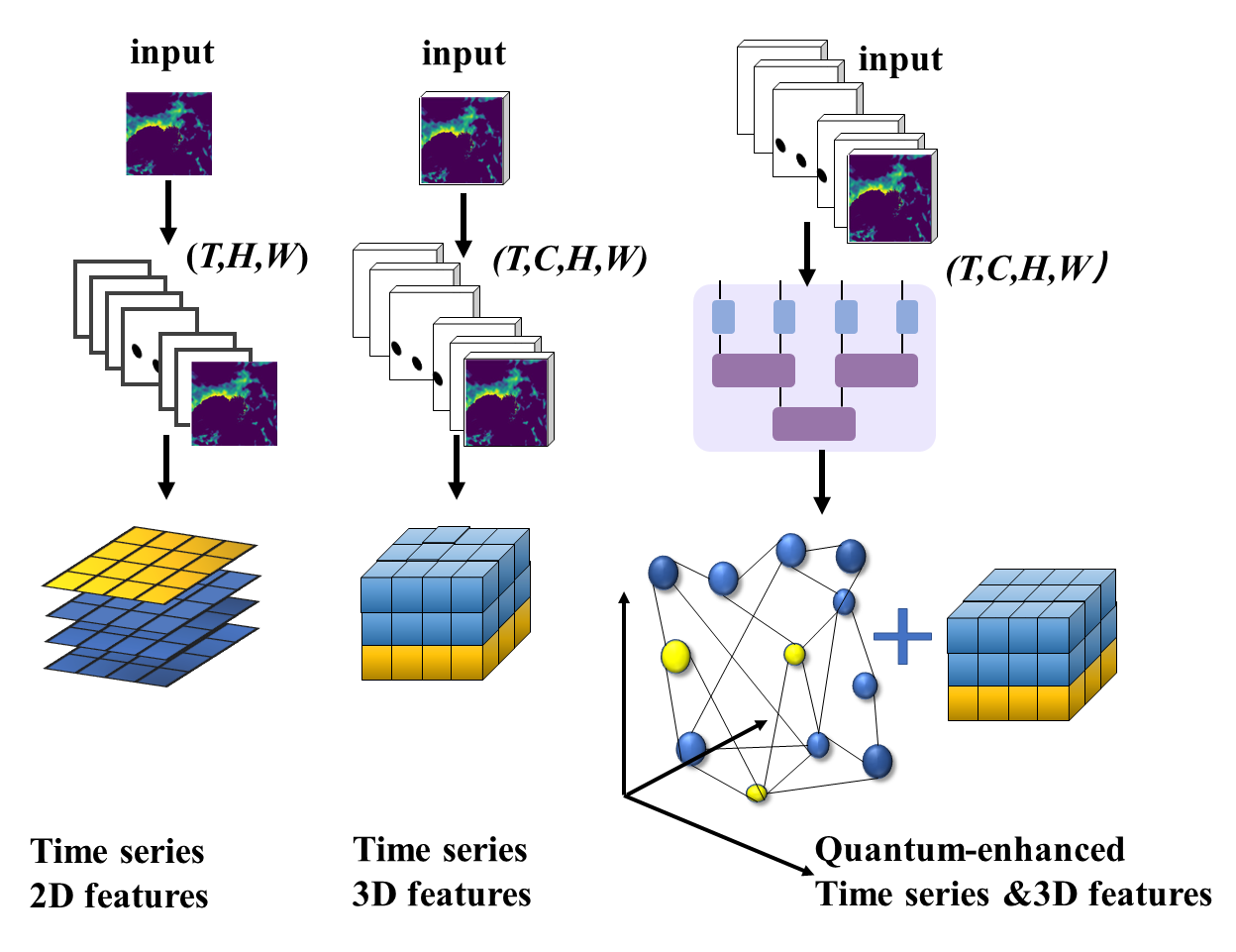}
  \caption{Comparison of 2D, 3D CNN and quantum-enhanced feature extraction.} 
  \label{fig1}
\end{figure}

The remainder of this paper is organized as follows. Section II reviews related work on classical spatiotemporal forecasting and hybrid quantum machine learning. Section III introduces the proposed QENO framework. Section IV reports the experimental results and ablation analysis. Section V concludes the paper and discusses future directions.


\section{Related Work}
\subsection{Classical Spatiotemporal Prediction Model}
\subsubsection{RNN-based Methods}
RNN–based spatiotemporal models excel at capturing temporal dependencies from historical data. By using gated units such as LSTM or GRU, they mitigate vanishing gradients and improve the stability and accuracy of short-term forecasts\cite{waqas2024critical}. A classic example is ConvLSTM\cite{shi2015convolutional}, which embeds convolution operations directly into both its input–state and state–state transitions. This design effectively learns local spatiotemporal correlations—such as those in precipitation radar echoes—but its inherently sequential structure limits its ability to model long-range dependencies, like the energy exchanges between distant cloud layers.
PredRNN\_plus\cite{wang2018predrnn++} tackles this “deep-time dilemma” by introducing the Causal LSTM and the Gradient Highway Unit (GHU). The Causal LSTM uses cascaded dual memory cells to better capture short-range dynamics, while the GHU provides skip connections that ease gradient flow through long sequences. Yet, because it still relies on local convolutions, it struggles to represent non-local interactions across the vertical structure of clouds.
PhyDNet\cite{guen2020disentangling} incorporates physical priors via its PhyCell module, which models simplified PDE dynamics in a latent space and delegates residual dynamics to a ConvLSTM branch. This hybrid approach enhances both predictability and interpretability for physical systems such as sea-surface temperature. However, the reliance on idealized PDE assumptions hampers its adaptability to the highly complex, non-local behaviors characteristic of three-dimensional cloud structures\cite{rasp2018deep}.

\subsubsection{CNN-Based Methods and attention mechanisms}
Convolutional neural networks (CNNs) augmented with attention mechanisms improve feature extraction by combining local convolutional filters with global attention weighting. For example, the Motion‐Aware Unit (MAU)\cite{chang2021mau} introduces a spatially guided attention module that aggregates historical states according to spatial correlations, thereby widening the temporal receptive field and improving motion modeling accuracy. However, because its attention weights rely on classical similarity metrics, MAU struggles to capture nonlinear, non‐local dependencies in high‐dimensional data such as cloud vertical profiles.
SimVP\cite{gao2022simvp} employs a pure CNN backbone—an encoder–transformer–decoder stack—that sidesteps recurrent bottlenecks by layering convolutions and Inception modules. While it delivers remarkable speed and accuracy on benchmarks like Moving MNIST, its convolutional operations remain inherently local and cannot capture long-range interactions between distant cloud layers. SimVP\_plus\cite{tan2025simvpv2} builds on this design by adding gated spatiotemporal attention and large-kernel convolutions to approximate extended dependencies and improve efficiency. However, it still operates under the classical locality assumption and thus cannot fundamentally extract truly non-local features.
Earthformer\cite{gao2022earthformer} advances cube attention by partitioning spatiotemporal data into local 3D cubes and using global context vectors to model long‐range dependencies. While it excels in precipitation nowcasting, its cube decomposition still adheres to a locality prior, and the attention mechanism incurs high computational costs for the high dimensionality of cloud fields. Finally, although details on TAU\cite{tan2023temporal} are scarce, its design similarly focuses on optimizing local feature extraction within a classical paradigm, leaving non‐local correlation modeling an open challenge. 

\subsection{Quantum-enhanced Machine Learning Model}
Quantum Neural Network (QNN) has demonstrated potential in image classification and molecular simulation. Former studies have attempted to extend QNN to time-series tasks, such as Quantum-LSTM\cite{chen2022quantum}. However, these methods suffer from two major limitations. One is that classical-to-quantum encodings (e.g., amplitude or angle encoding) often fail to preserve structure in high-dimensional inputs, undermining any quantum advantage\cite{wang2025limitations}. The other one is existing designs lack physical grounding and don’t leverage problem-specific topology, making it hard to relate entanglement patterns to real-world dynamics such as cloud evolution\cite{ranga2024quantum}.  
To overcome these issues, this paper introduces QENO, the first framework to align quantum entanglement patterns with the spatial topology of 3D cloud structures, ensuring that qubit correlations mirror actual cloud features. Also, it implements a fusion mechanism where quantum measurement outcomes adaptively gate the weights of a classical spatiotemporal backbone, enabling seamless information exchange between quantum and classical modules.

\begin{figure*}
  \centering 
  \includegraphics[width=1\textwidth]{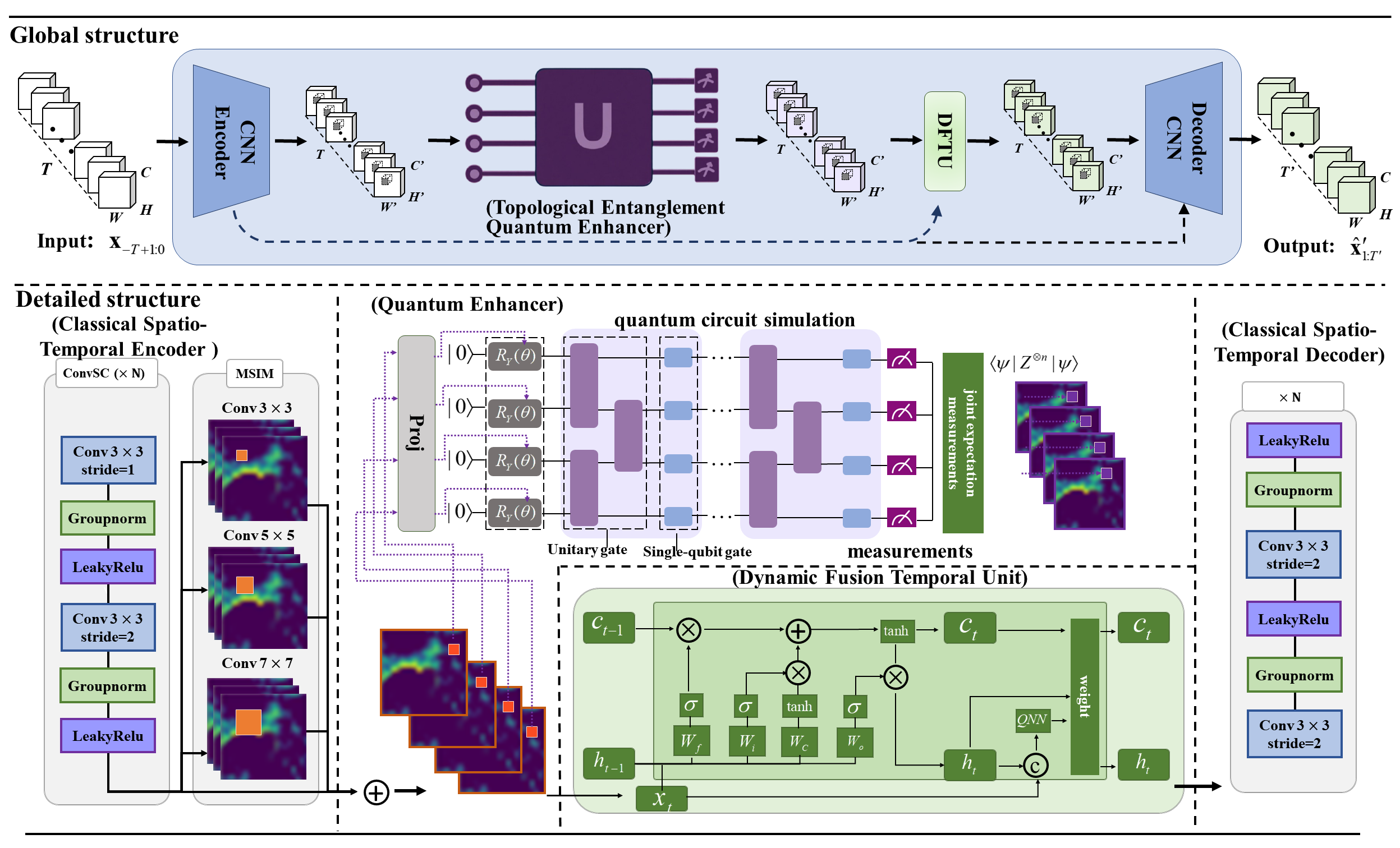} 
  \caption{The overall architecture of QENO} 
  \label{fig2} 
\end{figure*}

\section{The Proposed Quantum-Enhanced Method}
The QENO framework integrates classical and quantum processing in a four-stage pipeline: (1) a lightweight Classical Spatio-Temporal Encoder that extracts compact latent maps from past 3D cloud fields; (2) a Topological Entanglement Quantum Enhancer (TEQE) that maps latents onto a topology-mirroring qubit network and applies parameterized gates to generate non-local, multi-layer couplings; (3) a Dynamic Fusion Temporal Unit (DFTU) that gates a small classical recurrent cell using quantum measurement outcomes, enabling bidirectional interaction between quantum states and classical memory; and (4) a Classical Spatio-Temporal Decoder that upsamples and reconstructs the next 3D cloud field. The model ingests spatio-temporal inputs of shape \(T\times C\times H\times W\) (time steps, channels, height, width) and processes them through the four modules described below.

\subsection{Classical Spatial-Temporal Encoder}
The Classical Spatial-Temporal Encoder projects high-dimensional cloud fields into a compact latent space to reduce the computational burden of quantum encoding. Its architecture interleaves Convolutional Skip Connection (ConvSC) blocks and Multi-Scale Inception Modules (MSIM). ConvSC performs hierarchical downsampling via a learned stride generator, while MSIM applies parallel convolutions with kernels \(\{3\times3,5\times5,7\times7\}\) to capture both convective and advective structures. The encoder outputs a reduced-dimension tensor
\[
Z \in \mathbb{R}^{B\times T\times C'\times H'\times W'},\qquad
C'\ll C,\; H'\ll H,\; W'\ll W,
\]
which is consumed by the quantum enhancer.

\subsection{Topological Entanglement Quantum Enhancer}
In operational NWP systems cloud fields are often encoded as binary masks (cloud / clear-sky). We exploit this 0/1 parameterization to compress volumetric cues into a small qubit register via angle (or amplitude) encoding: quantum superposition enables a compact, simultaneous representation of many binary configurations, while a short sequence of parameterized single-qubit rotations and topology-aware entangling gates transforms local 0/1 contrasts into globally coherent quantum amplitudes. The entanglement graph is constructed to mirror key spatial patterns, so that the imposed CNOT/controlled-phase connectivity preferentially encodes physically meaningful cross-layer couplings.

Formally, let the encoder produce a pooled (flattened) latent vector \(v=\mathrm{vec}(Z)\) for each sample and time step. We map \(v\) to rotation phases \(\phi\) with a small linear layer followed by a sigmoid and scaling to \([0,\pi]\):
\begin{equation}\label{eq:phi}
\phi \;=\; \pi\cdot\sigma\bigl(W\,v + b\bigr) \,,
\end{equation}
where \(W,b\) are learnable parameters and \(\sigma(\cdot)\) denotes the sigmoid function.

Using angle encoding, the classical data are injected into the quantum register by applying single-qubit rotations to an initial ground state:
\begin{equation}\label{eq:encode}
\lvert\psi_0(\phi)\rangle \;=\; \bigotimes_{i=1}^{q} R_{y}(\phi_i)\,\lvert 0\rangle^{\otimes q},
\end{equation}
where \(q\) is the qubit count and \(R_y(\cdot)\) is the standard rotation operator. (Amplitude encoding can be used alternatively when qubit budget allows; here we focus on angle encoding for its simplicity and lower circuit depth.).

State evolution is realized by stacking \(L\) parameterized layers, each composed of local rotations and topology-defined entangling unitaries. One layer updates the state as
\begin{equation}\label{eq:layer_evol}
\lvert\psi_{\ell}\rangle 
\;=\; U_{\mathrm{ent}}(E)\;U_{\mathrm{RZ}}(\phi^{(\ell)})\;U_{\mathrm{RY}}(\phi^{(\ell)})\;U_{\mathrm{RX}}(\phi^{(\ell)})\;\lvert\psi_{\ell-1}\rangle,
\end{equation}
where \(U_{\mathrm{ent}}(E)\) is the entangling unitary defined by adjacency \(E\) (e.g., CNOTs along ring/chain edges) and \(U_{\mathrm{RX/RY/RZ}}\) denote layerwise single-qubit rotations parameterized by the corresponding phase subsets \(\phi^{(\ell)}\).

After \(L\) layers we extract classical readouts via expectation values of selected observables (Pauli strings). Let \(\{P_j\}_{j=1}^m\) denote the chosen measurement operators (the special case \(P_j=Z^{\otimes q}\) is a global parity readout). The measurement vector \(m\in\mathbb{R}^m\) is
\begin{equation}\label{eq:measure}
m_j \;=\; \bigl\langle \psi_L \bigr| P_j \bigl| \psi_L \bigr\rangle,\qquad j=1,\dots,m.
\end{equation}

Finally, the measurement vector is mapped back to a spatial feature tensor via a small decoder head and reshaping:
\begin{equation}\label{eq:map_back}
Q_t \;=\; \mathrm{Reshape}\bigl(\mathrm{FC}(m)\bigr)
\;\in\; \mathbb{R}^{B\times C'\times H'\times W'},
\end{equation}
producing quantum-enhanced feature maps \(Q_t\) that retain spatial alignment with the encoder latent grid. These maps are concatenated or fused with classical latents and passed to the Dynamic Fusion Temporal Unit (DFTU) for temporal integration and subsequent decoding.

\vspace{1ex}\noindent\textbf{Implementation notes.} In practice we keep \(q\) and \(L\) small (e.g., \(q\lesssim 8\), modest \(L\)) to limit simulator cost; the choice of measurement operators \(\{P_j\}\) balances expressivity and readout dimensionality. The topology selector \(E\) may be rule-based (simple motif detectors) or learned; both variants are differentiable end-to-end when the selector is implemented as a soft adjacency head.

\subsection{Dynamic Fusion Temporal Unit}
The DFTU fuses classical recurrent dynamics with quantum-enhanced features. Let the classical LSTM cell produce intermediate states
\begin{equation}\label{eq:lstm}
(h_t^{\mathrm{cls}},c_t^{\mathrm{cls}})=\mathrm{LSTMCell}\bigl(\mathrm{flatten}(z_t),\,(h_{t-1},c_{t-1})\bigr).
\end{equation}
We pool and linearly transform the quantum map \(Q_t\) into a vector and compute a trainable gate
\begin{equation}\label{eq:gate}
g = \sigma\bigl(W_g\,[\,h_t^{\mathrm{cls}};\;\mathrm{pool}(Q_t)\,]+b_g\bigr),
\end{equation}
which decomposes into four coefficients \((g_0,g_1,g_2,g_3)\). The fused hidden and cell states are then updated as follows:
\begin{equation}\label{eq:fusion}
\begin{aligned}
h_t &\leftarrow g_0\odot h_t^{\mathrm{cls}} \;+\; g_1\odot \mathrm{FC}(Q_t),\\
c_t &\leftarrow g_2\odot c_t^{\mathrm{cls}} \;+\; g_3\odot \mathrm{FC}(Q_t).
\end{aligned}
\end{equation}
The enhanced latent \(z_t^{\mathrm{enh}}\) concatenates the original \(z_t\) and a reshaped \(h_t\) to feed the decoder.

\subsection{Classical Spatial-Temporal Decoder}
The Classical Spatial-Temporal Decoder reconstructs high-resolution volumetric cloud forecasts from \(\{z_t^{\mathrm{enh}}\}\). It uses transposed convolutional layers with encoder skip-connections and a final 3D convolutional post-processing stage to refine boundaries and correct residual errors, producing \(\hat{X}\in\mathbb{R}^{B\times T_{\mathrm{out}}\times C\times H\times W}\).

\subsection{Algorithmic Summary}
Algorithm~\ref{alg:qeno} summarizes the end-to-end training and inference workflow (matching Figure~2). Pseudocode highlights: (i) encoder \(\rightarrow\) topology selector; (ii) per-frame Topological Entangler producing spatial quantum maps \(Q_t\); (iii) DFTU fusion for temporal integration; (iv) decoder reconstruction.

\begin{algorithm}[ht]
    \scriptsize  
    \caption{Concise QENO data-flow}
    \label{alg:qeno}
    \KwIn{Input $X$ of shape B x T\_in x C x H x W}
    \KwOut{Forecast $\hat{Y}$ of shape B x T\_out x C x H x W}
    
    1: \quad Z := Encoder(X) \tcp*{ConvSC + MSIM $\rightarrow$ compressed latent Z}
    2: \quad E := TopologySelector(Z) \tcp*{rule-based or learned adjacency}
    3: \quad initialize h\_0, c\_0 := 0\;
    4: \quad \For{$t=1$ \KwTo $T_{\text{in}}$}{
        \quad\quad z\_t := Z[:,t,...]\;
        \quad\quad p\_t := Proj(z\_t) \tcp*{pool + linear projection for quantum input}
        \quad\quad psi\_t := QuantumCircuitSimulation(p\_t, E, L) \tcp*{simulate shallow circuit}
        \quad\quad Q\_t := Measure(psi\_t) \tcp*{joint expectation $\to$ spatial map}
        \quad\quad (z\_t\_enh, h\_t, c\_t) := DFTU(z\_t, h\_{t-1}, c\_{t-1}, Q\_t)\;
    }
    5: \quad \(\hat{Y}\) := Decoder(stack of z\_t\_enh for t=1..T\_in)\;
    6: \quad loss := Loss(\(\hat{Y}\), Y\_gt); backpropagate and update parameters\;
\end{algorithm}

\section{Experiment and Results}
\subsection{Dataset Desciption and Baseline Models}

The experiments employ 3D cloud fields generated by the CMA‐MESO, an operational regional NWP system, featuring 3 km spatial resolution, 3‐hour temporal resolution, and a cropped domain of \(64\times64\) grid points across 42 vertical levels. All model training and inference were performed on an NVIDIA Tesla V100 GPU with CUDA acceleration. For the quantum component, we utilized the torchquantum framework to simulate parameterized quantum circuits and integrate their outputs with classical spatiotemporal models. Each model was trained for 100 epochs using a batch size of 8 and an initial learning rate of 0.001, optimized via the mean squared error (MSE) loss. Forecasts of the next two time steps (\(T+1\) and \(T+2\)) were generated from the preceding five-frame sequence (\(T-4\) to \(T\)).
To benchmark our approach, we compared QENO against eight state‐of‐the‐art baselines: ConvLSTM, PhyDNet, MAU, SimVP, SimVP\_Plus, Earthformer, TAU, and PredRNN\_Plus. This comprehensive evaluation quantifies the benefit of quantum‐enhanced feature extraction in 3D cloud forecasting.  
The quantitative analysis carried out objective comparative experiments on the regression performance and spatio-temporal performance of the model to evaluate the performance of QENO. In addition, we also conducted ablation experiments to demonstrate the effectiveness of our Quantum Enhanced Mid module and Quantum Enhanced Decoder module.

\begin{figure*}
  \centering 
  \includegraphics[width=0.8\textwidth]{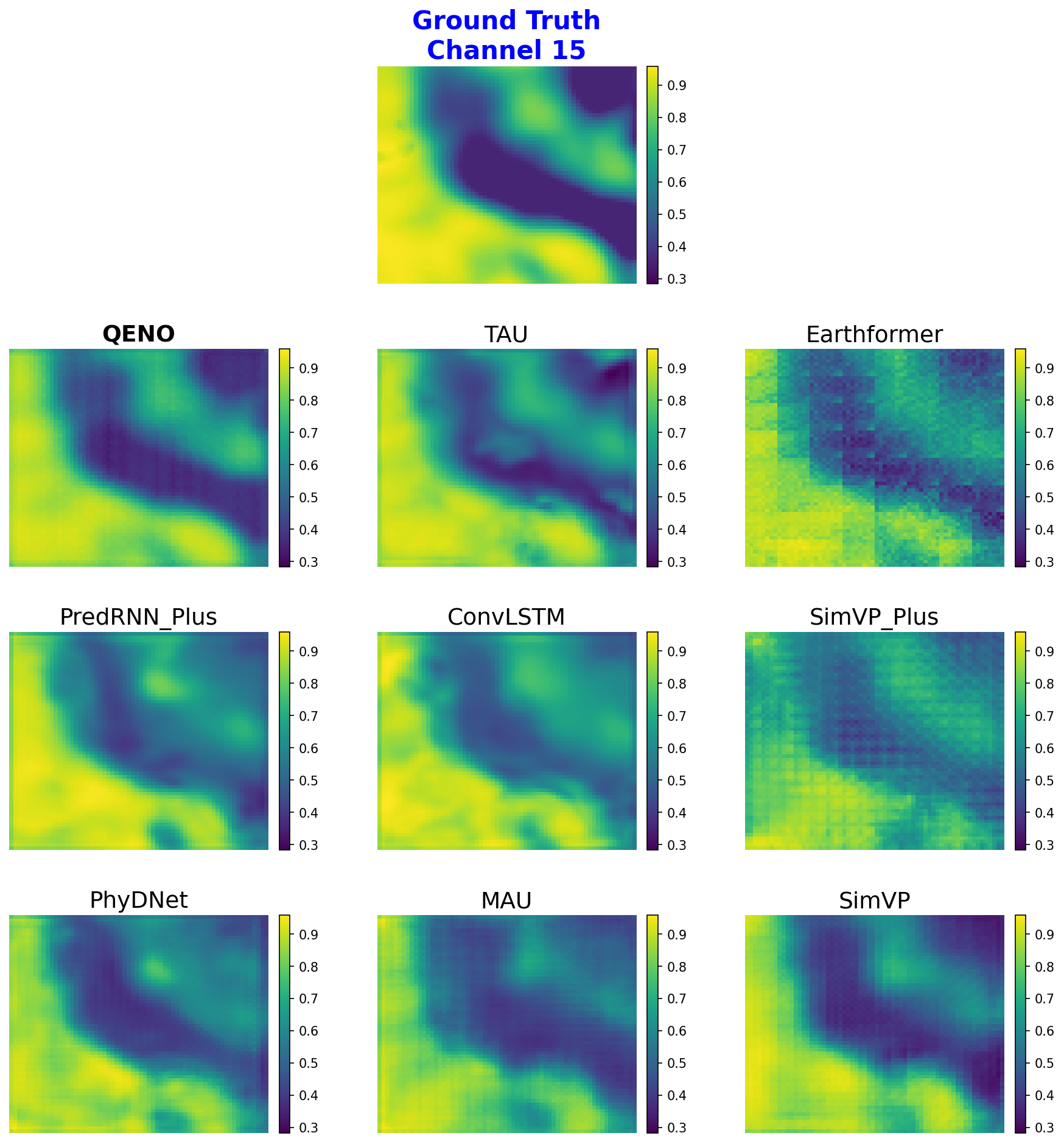}
  \caption{Visualization of 3D cloud forecasts for Channel 15 from the CMA‐MESO dataset. Top: Ground truth. Below: model predictions by QENO, TAU, Earthformer, PredRNN\_Plus, ConvLSTM, SimVP\_Plus, PhyDNet, MAU, and SimVP. This layout illustrates each model’s capability to reproduce fine‐scale cloud patterns and overall structural fidelity.} 
  \label{fig3}
\end{figure*}

\subsection{Analysis of Comparative Experiment}

\subsubsection{Quantitative Analysis}
Four commonly used evaluation metrics are employed to assess the prediction, namely the Mean Squared Error (MSE), Mean Absolute Error (MAE), Root Mean Squared Error (RMSE), and Structural Similarity Index (SSIM). Lower MSE, MAE, RMSE, and higher SSIM indicate better predictions. 

\begin{table}[htbp]
    \centering
    \caption{Performance comparison of different models on the dataset.}
    \scriptsize  
    \setlength{\tabcolsep}{7pt}  
    \begin{tabularx}{0.65\linewidth}{lXXXXX}
        \toprule
        Model & MSE & MAE & RMSE & SSIM & Params \\
        \midrule
        QENO & \textbf{0.2038} & \textbf{0.2553} & \textbf{0.4514} & \textbf{0.6291} & \textbf{0.03M}\\
        TAU & 0.2372 & 0.2757 & 0.4871 & 0.5063 & 0.44M\\
        Earthformer & 0.2566 & 0.3120 & 0.5065 & 0.4637 & 6.35M \\
        PredRNN\_Plus & 0.2726 & 0.3079 & 0.5221 & 0.4586 & 2.50M\\
        ConvLSTM & 0.2883 & 0.3100 & 0.5370 & 0.4594 & 0.60M\\
        SimVP\_Plus & 0.3477 & 0.3470 & 0.5897 & 0.3965 & 0.30M \\
        PhyDNet & 0.3025 & 0.3162 & 0.5500 & 0.4187 & 0.55M\\
        MAU & 0.3204 & 0.3458 & 0.5660 & 0.4250 & 2.64M \\        SimVP & 0.3940 & 0.3610 & 0.6277 & 0.3194 & 0.53M\\
        \bottomrule
    \end{tabularx}
    \label{tab1}
\end{table}

In Table~\ref{tab1}, QENO performs excellently on all regression metrics. Especially regarding MSE (0.2038), MAE (0.2553), and RMSE (0.4515), its performance is more accurate compared to other competing models. The SSIM value (0.6292) of QENO is also significantly higher than that of other models, indicating that it has a significant advantage in maintaining structural similarity. Particularly when compared with SimVP (MSE 0.3940, RMSE 0.6277) and SimVP\_Plus (MSE 0.3477, RMSE 0.5897), the MSE and RMSE of QENO are significantly lower, demonstrating its higher accuracy in spatio-temporal structure prediction. Notably, QENO achieves these gains while using the smallest parameter budget (0.03M), substantially fewer parameters than competing models. This parameter efficiency directly reflects the compact qubit-based topological encoding: by mapping volumetric cloud motifs to sparse entanglement graphs, the Topological Entanglement Quantum Enhancer captures high-order, non-local couplings with far fewer learnable classical weights—yielding strong representational power at a much lower parameter cost.

Three evaluation metrics, including the Critical Success Index (CSI), the Probability of Detection (POD), and the Heidecker Skill Score (HSS), are adopted to evaluate the performance of the model on the test set. These metrics quantitatively analyze the predictive ability of the model from different perspectives and can comprehensively reflect the performance of the model under different cloud field intensities. The CSI value ranges from 0 to 1. The higher the value, the more accurate the prediction. The POD value also ranges from 0 to 1. The larger the value, the higher the sensitivity of the model to actual cloud field events. The HSS value ranges from -1 to 1. The larger the value, the better the overall predictive performance. To evaluate the performance of the model under different cloud field intensities, we set different thresholds. For each threshold, we binarize the predicted and actual cloud field data values: if the predicted or actual cloud field value is greater than or equal to the threshold, it is marked as 1; otherwise, it is marked as 0.

\[
B_{\tau}\left(x_{i j}\right)=
\left\{
\begin{array}{ll}
0, & \text{if } 0 \leq x_{i j} < \tau \\
1, & \text{if } x_{i j} \geq \tau
\end{array}
\right.
\]

where \(x_{ij}\) represents the predicted precipitation value or the true cloud field value at grid point \((i, j)\), and \(B_{\tau}(x_{ij})\) represents the conversion of the cloud field value into corresponding binary data based on the cloud field threshold \(\tau\). By using Equation 10, we transform the problem into a binary classification task. We can calculate the value of each evaluation metric according to the formula.

\begin{table*}[htbp]
    \centering
    \caption{Summary of Critical Success Index (CSI), Heidke Skill Score (HSS), and Probability of Detection (POD) for each forecasting model evaluated at multiple threshold levels.}
    \begin{subtable}[t]{\linewidth}
        \centering
        \caption{CSI values for each forecasting model.}
        \scriptsize  
        \setlength\tabcolsep{4.2pt}  
        \begin{tabularx}{\linewidth}{lccccc}
            \toprule
            Model & CSI-0.1 & CSI-0.3 & CSI-0.5 & CSI-0.7 & CSI-0.8 \\
            \midrule
            QENO & \textbf{0.8764($\uparrow$22\%)} & \textbf{0.8701($\uparrow$19\%)} & \textbf{0.8565($\uparrow$20\%)} & \textbf{0.8383($\uparrow$23\%)} & \textbf{0.8275($\uparrow$23\%)} \\
            TAU & 0.7023 & 0.7074 & 0.6991 & \underline{0.6812} & \underline{0.6678} \\
            Earthformer & 0.6874 & 0.7092 & 0.6978 & 0.6685 & 0.6485 \\
            PredRNN\_Plus & 0.6837 & 0.6856 & 0.6746 & 0.6538 & 0.6396 \\
            SimVP\_Plus & \underline{0.7158} & \underline{0.7287} & \underline{0.7124} & 0.6741 & 0.6468 \\
            PhyDNet & 0.7015 & 0.6850 & 0.6474 & 0.6009 & 0.5755 \\
            MAU & 0.7041 & 0.7050 & 0.6840 & 0.6482 & 0.6238 \\
            SimVP & 0.6930 & 0.6876 & 0.6635 & 0.6249 & 0.5999 \\
            ConvLSTM & 0.6541 & 0.6662 & 0.6615 & 0.6455 & 0.6333 \\
            \bottomrule
        \end{tabularx}        
        \label{tab2a}
    \end{subtable}
    
    \vspace{2pt}
    
    \begin{subtable}[t]{\linewidth} 
        \centering
        \scriptsize  
        \caption{HSS values for each forecasting model.}
        \setlength\tabcolsep{4.2pt}  
        \begin{tabularx}{\linewidth}{lccccc}
            \toprule
            Model & HSS-0.1 & HSS-0.3 & HSS-0.5 & HSS-0.7 & HSS-0.8 \\
            \midrule
            QENO & \textbf{0.9131($\uparrow$17\%)} & \textbf{0.9109($\uparrow$14\%)} & \textbf{0.9035($\uparrow$14\%)} & \textbf{0.8931($\uparrow$16\%)} & \textbf{0.8867($\uparrow$16\%)} \\
            TAU & 0.7619 & 0.7761 & 0.7770 & \underline{0.7692} & \underline{0.7613} \\
            Earthformer & 0.7497 & 0.7815 & 0.7802 & 0.7628 & 0.7494 \\
            PredRNN\_Plus & 0.7465 & 0.7583 & 0.7571 & 0.7467 & 0.7379 \\
            SimVP\_Plus & \underline{0.7764} & \underline{0.7978} & \underline{0.7918} & 0.7670 & 0.7479 \\
            PhyDNet & 0.7671 & 0.7628 & 0.7388 & 0.7060 & 0.6871 \\
            MAU & 0.7680 & 0.7788 & 0.7690 & 0.7462 & 0.7289 \\
            SimVP & 0.7571 & 0.7628 & 0.7506 & 0.7252 & 0.7070 \\
            ConvLSTM & 0.7149 & 0.7389 & 0.7445 & 0.7391 & 0.7325 \\
            \bottomrule
        \end{tabularx}
        \label{tab2b}
    \end{subtable}
    
    \vspace{2pt}
    
    \begin{subtable}[t]{\linewidth} 
        \centering
        \caption{POD values for each forecasting model.}
        \scriptsize  
        \setlength\tabcolsep{5.5pt}  
        \begin{tabularx}{\linewidth}{lccccc}
            \toprule
            Model & POD-0.1 & POD-0.3 & POD-0.5 & POD-0.7 & POD-0.8 \\
            \midrule
            QENO & \textbf{0.9379($\uparrow$2\%)} & \textbf{0.9353($\uparrow$5\%)} & \textbf{0.9307($\uparrow$8\%)} & \textbf{0.9199($\uparrow$11\%)} & \textbf{0.9111($\uparrow$13\%)} \\
            TAU & \underline{0.9171} & \underline{0.8905} & \underline{0.8588} & \underline{0.8219} & \underline{0.7996} \\
            Earthformer & 0.8819 & 0.8390 & 0.7933 & 0.7429 & 0.7153 \\
            PredRNN\_Plus & 0.8752 & 0.8480 & 0.8167 & 0.7800 & 0.7585 \\
            SimVP\_Plus & 0.9004 & 0.8622 & 0.8153 & 0.7561 & 0.7203 \\
            PhyDNet & 0.8429 & 0.7869 & 0.7245 & 0.6609 & 0.6287 \\
            MAU & 0.8625 & 0.8220 & 0.7736 & 0.7172 & 0.6842 \\
            SimVP & 0.8609 & 0.8157 & 0.7639 & 0.7049 & 0.6715 \\
            ConvLSTM & 0.8819 & 0.8523 & 0.8177 & 0.7770 & 0.7532 \\
            \bottomrule
        \end{tabularx}        
        \label{tab2c}
    \end{subtable}
    \label{tab2}
\end{table*}

\begin{figure*}
  \centering 
  \includegraphics[width=1\textwidth]{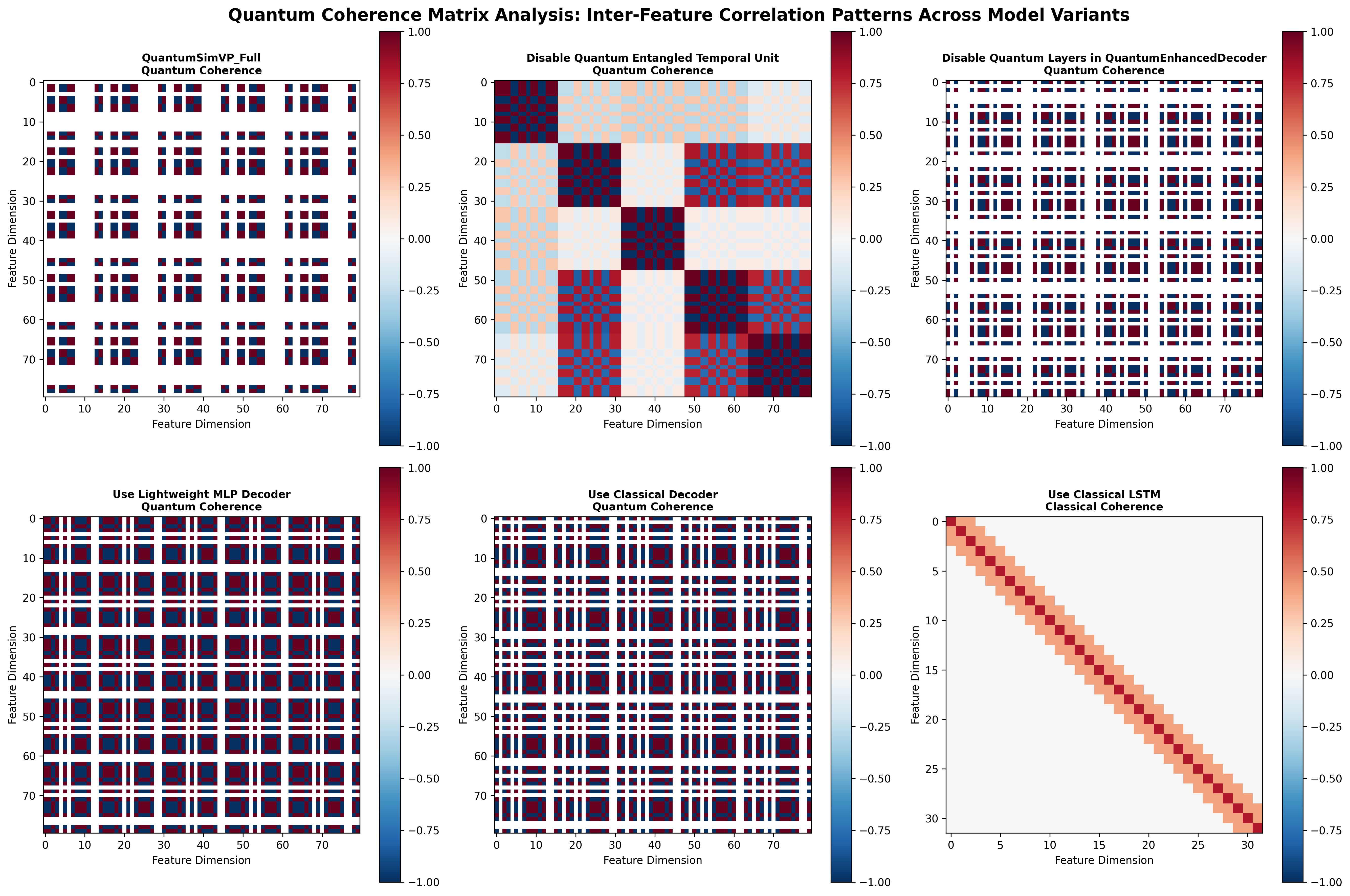}
  \caption{Quantum coherence matrix analysis: inter-feature correlation patterns across model variants. Panels (a)--(f) show pairwise correlation (coherence) of latent feature activations computed over the validation set for six model variants: (a) QENO\_Full (full quantum-enhanced model), (b) QENO without the Dynamic Fusion Temporal Unit (DFTU disabled), (c) QENO with quantum layers removed from the decoder, (d) QENO with a lightweight MLP decoder, (e) QENO with a classical convolutional decoder, and (f) a purely classical LSTM baseline. Color encodes Pearson-style correlation (range \([-1,1]\)); strong off-diagonal structure indicates pronounced non-local feature coherence. The full QENO model (a) exhibits large-scale block structure and prominent off-diagonal correlations, whereas ablated or classical variants (b)--(f) progressively lose these organized long-range patterns and revert toward local/diagonal correlation.}
  \label{fig4}
\end{figure*}

In Table~\ref{tab2}, the suffixes (e.g., 0.1, 0.3, 0.5, 0.7, 0.8) denote the cloud‐mask threshold applied to both forecasts and observations—so, for instance, “CSI-0.1” evaluates the CSI when the cloud mask exceeds 0.1. Especially for CSI-0.1 (0.8764), CSI-0.3 (0.8700), and CSI-0.5 (0.8565), the CSI values were higher than those of all the comparison models, indicating that it has stronger accuracy in cloud field structure prediction. Compared with SimVP+ (CSI-0.1 0.7158) and SimVP (CSI-0.1 0.6930), QENO showed a higher prediction success rate, which was particularly prominent under high thresholds.
HSS (Heidecker Skill Score): In terms of the HSS metric, QENO also demonstrated powerful performance, especially for HSS-0.1 (0.9131) and HSS-0.3 (0.9109). In contrast, SimVP+ (HSS-0.1 0.7764) and SimVP (HSS-0.1 0.7571) had weaker overall performance, indicating that QENO has stronger overall prediction ability under different thresholds. QENO also performed better than other models at HSS-0.8 (0.8867), especially under relatively high thresholds, indicating that it has higher sensitivity and accuracy in predicting high-intensity cloud field events.
POD (Probability of Detection): QENO also had an advantage in POD, especially for POD-0.1 (0.9379) and POD-0.3 (0.9353). Its POD values were significantly higher than other models, indicating that it has a stronger ability to capture actual cloud field events. SimVP+ (POD-0.1 0.9004) and SimVP (POD-0.1 0.8609) had lower values in the POD metric, indicating that their detection ability was weaker under relatively high thresholds.

\subsubsection{Qualitative Analysis}
As shown in Figure~\ref{fig3}, QENO produces visibly sharper and more structurally faithful forecasts on Channel~15 than competing methods: its predictions preserve fine-scale details and cloud boundaries that other models fail to restore. Baselines such as TAU (SSIM~0.4443), Earthformer (SSIM~0.3922) and PredRNN\_Plus (SSIM~0.3871) capture broad features but exhibit blurring and distortion, particularly near edges and in structurally complex regions; SimVP and SimVP\_Plus perform worse still. These qualitative differences mirror the quantitative gains: QENO's higher SSIM and improved detail fidelity indicate that the topology-aware quantum enhancement and DFTU fusion better capture non-local, multi-scale spatio-temporal dependencies, yielding more accurate 3D cloud-structure forecasts.

\subsection{Ablation Studies}

\begin{table}[htbp]
  \centering
  \caption{Impact of ablation studies on training‐phase performance metrics, QENO minus QEDecoder, QENO minus QEMid, and SimVP (which is QENO minus both  QEDecoder and QEMid)}
  \small
  \setlength\tabcolsep{3pt}
  \begin{tabularx}{0.8\linewidth}{lcccc}
    \toprule
    Model & MSE & MAE & RMSE & SSIM\\
    \midrule
    QENO & \textbf{0.201300} & \textbf{0.249777} & \textbf{0.448665} & \textbf{0.680145}\\
    QENO-QEDecoder& 0.292149 & 0.284594 & 0.540508 & 0.673665\\
    QENO-QEMid  & 0.899545 & 0.638966 & 0.948443 & 0.193796\\
    SimVP & 0.48236 & 0.390819 & 0.694521 & 0.286895\\
    \bottomrule
  \end{tabularx}
  
  \label{tab3}
\end{table}

To verify the contribution of each component in QENO, we conducted ablation experiments that individually disable the Quantum-Enhanced Middle module (QEMid) and the Quantum-Enhanced Decoder module (QEDecoder), and compared results against the SimVP baseline; the results are reported in Table~\ref{tab3}. Disabling different modules has a substantial effect on performance, with the full QENO model achieving the best scores across all metrics (MSE 0.2013, MAE 0.2498, RMSE 0.4487) and a high SSIM (0.6801), indicating superior accuracy and structural fidelity.

\subsubsection{QENO-QEDecoder}
In the model after turning off the Quantum-Enhanced Decoder module, although the MSE (0.2921), MAE (0.2846), and RMSE (0.5405) increase compared with QENO, they are still significantly better than those of SimVP (MSE 0.4824, RMSE 0.6945). However, the SSIM (0.6737) slightly decreases, indicating that the Quantum-Enhanced Decoder module plays a certain role in maintaining details and structure. 
\subsubsection{QENO-QEMid}When the Quantum-Enhanced Intermediate Module is turned off, the model performance drops significantly. In particular, the MSE (0.8995) and RMSE (0.9484) increase significantly, indicating that the decoder module is crucial for restoring the accuracy of high-dimensional predictions. At the same time, the SSIM (0.1938) drops substantially, suggesting that the absence of the intermediate module leads to a significant loss of structural fidelity. SimVP: As a benchmark model, SimVP performs relatively poorly. Both the MSE (0.4824) and RMSE (0.6945) are higher than the corresponding values of QENO, and the SSIM is also lower (0.2869). This indicates that SimVP has obvious accuracy deficiencies in spatio-temporal prediction modeling, especially in the prediction task of complex cloud field structure . 
Finally, Figure~4 supports these findings at the representational level: the full QENO exhibits pronounced block-like off-diagonal coherence—evidence of organized, non-local inter-feature coupling—whereas ablating quantum or fusion components progressively degrades these off-diagonal patterns into more local or fragmented correlations; the classical LSTM baseline remains largely diagonal, indicating primarily self-correlations.

In summary, the results of the ablation experiments verify the importance of the two quantum-enhanced modules in QENO. Turning off any one of these modules will lead to a significant decline in model performance. In particular, turning off the QuantumEnhancedMid module results in a serious loss of prediction accuracy and structural fidelity. This further demonstrates the crucial role of the quantum-enhanced modules in improving the prediction ability of QENO, especially their advantages in capturing non-local dependencies and high-dimensional feature representations.

\section{Conclusion}
In this work, we introduce QENO, a hybrid quantum–classical framework for spatiotemporal forecasting of 3D cloud fields. By integrating a topology-aware entanglement layer, dynamic fusion gating, and an efficient quantum–classical interface, QENO overcomes the inherent limitations of classical CNN- and RNN-based predictors in capturing non-local dependencies and high-dimensional feature interactions. Experimental evaluations on operational 3D cloud datasets demonstrate that the quantum-enhanced components substantially elevate forecast accuracy—particularly under complex, multiscale cloud evolution scenarios—outperforming leading time-series architectures such as ConvLSTM and SimVP. QENO’s superior ability to model cross-layer and long-range spatiotemporal correlations marks a significant advance in high-fidelity cloud prediction. Future work will focus on reducing the computational overhead of quantum modules and extending the methodology to additional meteorological forecasting tasks, thereby charting new pathways for the practical deployment of quantum computing in Earth system modeling.

\bibliographystyle{IEEEtran}
\bibliography{Reference}

\end{document}